\documentclass[lettersize,journal]{IEEEtran}
\usepackage{amsmath,amsfonts}
\usepackage{algorithmic}
\usepackage{algorithm}
\usepackage{array}
\usepackage[caption=false,font=normalsize,labelfont=sf,textfont=sf]{subfig}
\usepackage{textcomp}
\usepackage{stfloats}
\usepackage{url}
\usepackage{verbatim}
\usepackage{graphicx}
\usepackage{cite}
\usepackage[bookmarks=true,pdfencoding=auto,breaklinks,colorlinks,hypertexnames=false]{hyperref}

\usepackage{amsfonts}
\usepackage{graphicx}
\usepackage{booktabs}
\usepackage{multirow}
\usepackage{multicol}

\usepackage[flushleft]{threeparttable}
\newcommand{\ours}{BYE}

\begin{document}

\title{BYE: Build Your Encoder with One Sequence of Exploration Data for Long-Term Dynamic Scene Understanding}

\author{Chenguang Huang, Shengchao Yan, and Wolfram Burgard,~\IEEEmembership{Fellow,~IEEE,}
        % <-this % stops a space
\thanks{Chenguang Huang and Shengchao Yan are with the Department of Computer Science, University of Freiburg, 79110 Freiburg, Germany (e-mail: \href{mailto:huang@informatik.uni-freiburg.de}{huang@informatik.uni-freiburg.de};\href{mailto:yan@informatik.uni-freiburg.de}{yan@informatik.uni-freiburg.de})}
\thanks{Wolfram Burgard is with the Department of Engineering, University of
Technology Nuremberg, 90443 Nuremberg, Bavaria, Germany (e-mail: \href{mailto:wolfram.burgard@utn.de}{wolfram.burgard@utn.de}).}% <-this % stops a space
}



\maketitle

\begin{abstract}
Dynamic scene understanding remains a persistent challenge in robotic applications. Early dynamic mapping methods focused on mitigating the negative influence of short-term dynamic objects on camera motion estimation by masking or tracking specific categories, which often fall short in adapting to long-term scene changes. Recent efforts address object association in long-term dynamic environments using neural networks trained on synthetic datasets, but they still rely on predefined object shapes and categories. Other methods incorporate visual, geometric, or semantic heuristics for the association but often lack robustness. In this work, we introduce \ours{}, a class-agnostic, per-scene point cloud encoder that removes the need for predefined categories, shape priors, or extensive association datasets. Trained on only a single sequence of exploration data, \ours{} can efficiently perform object association in dynamically changing scenes. We further propose an ensembling scheme combining the semantic strengths of Vision Language Models (VLMs) with the scene-specific expertise of \ours{}, achieving a 7\% improvement and a 95\% success rate in object association tasks. Code and dataset are available at \href{https://byencoder.github.io}{https://byencoder.github.io}.

% the idea of training a per-scene encoder
% the idea is similar to treating a neural network as part of the map?
% can we treat an encoder specializing in the contents in one environment?
% previous methods approach this like nerf or gaussian splitting, but none of them tried to represent the long-term dynamic scenes as a network. In this work, we make a first attempt to approach this by training a point cloud encoder and generate latent embeddings for 

% if possible, highlight spatio-temporal navigation and 3D prompt navigation
\end{abstract}

\begin{IEEEkeywords}
Representation Learning, Dynamic Mapping, Robot Navigation, Scene Understanding
\end{IEEEkeywords}

\section{Introduction}

\IEEEPARstart{I}{nteracting} with the physical environment is inherently dynamic. On one hand, numerous objects in our daily lives—such as people, animals, vehicles, and machines—are constantly in motion, requiring us to reason, react, avoid, maneuver, and operate while processing information instantly. On the other hand, beyond our immediate sight, the environment itself continues to evolve: objects are moved, drawers opened, screens turned off, and containers refilled. As we revisit or explore, we continuously associate new observations with past experiences, updating our knowledge base. This raises the question: can a robot similarly associate new observations with its existing knowledge base?

\begin{figure}[t]
\centering
\includegraphics[width=0.49\textwidth]{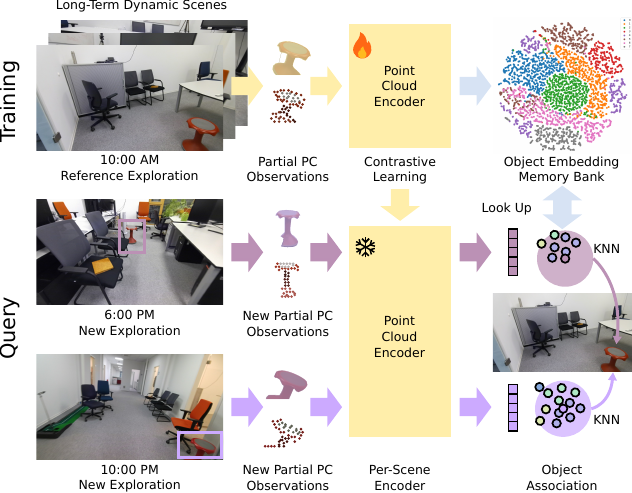}
\caption{BYE enables reliable and robust object association in long-term dynamic scenes where object relocation exists by training a per-scene encoder on one sequence of exploration data, without the need for large synthetic datasets, category assumptions, and shape priors. The object embedding memory bank stores the latent embeddings for all partial point cloud observations in the reference exploration data where each point represents a partial observation and each color represents an instance. The contrastive learning process trains the encoder to gather observations from the same instance while repelling observations from different instances.}
\label{fig:cover_lady}
\vspace{-1em}
\end{figure}

Despite the recent spike of robot generalist policy research~\cite{o2024open} which aims at training a reactive robot policy that handles dynamics through learning from a plethora of data, the majority of the robotic systems still rely on a knowledge base of the environment to operate within a certain scope. Such a knowledge base is usually a scene representation, namely, a map. Previous dynamic mapping techniques focused on reducing the impact of in-view  dynamic objects by removing or tracking certain classes of semantic masks during camera pose estimation~\cite{runz2017co,bescos2018dynaslam,yu2018ds,xu2019mid}, which struggles to handle long-term dynamic scenes. Later approaches to long-term dynamic scene understanding used visual, geometric, or semantic clues as heuristics for associations between scenes~\cite{adam2022objects,adam2023has}. However, these methods require perfect registration of two scenes. Recently, researchers proposed to train an encoder-decoder relying on synthetic datasets to perform scene object association, registration, and reconstruction across object location changes~\cite{zhu2024living}. However, the method still depends on a predefined set of object categories and shapes in ShapeNet~\cite{chang2015shapenet}.

To this end, we introduce Build Your Encoder (\ours{}), a class-agnostic per-scene point cloud encoder that removes the need for predefined categories, shape priors, or large association datasets. It only requires a sequence of exploration data to train and can generalize the association ability to long-term dynamic changes in the environment. \ours{} highly resembles a human's ability to recall objects and their past locations when seeing objects with similar shapes and appearances in new places. The idea of \ours{} stems from the work LangSplat~\cite{qin2024langsplat} which trains a per-scene autoencoder to encode CLIP features of all images collected in the same scene into a low dimensional latent space to accelerate the optimization of Gaussian Splatting~\cite{kerbl2023gaussiansplatting}. In this work, our major contributions are:

\begin{enumerate}
    \item We introduce \ours{}, a novel pipeline to train a per-scene point cloud encoder for object association in long-term dynamic environments. The training pipeline eliminates the need for predefined categories, large synthetic datasets, and any shape priors, but only requires one sequence of exploration data in the scene.
    \item We propose to construct an object memory bank with \ours{} encoder and all partial point cloud observations in the exploration data for object association in dynamic scenes, resembling human memory of past experiences.
    \item We propose an ensembling scheme that leverages the semantic strengths of VLMs alongside the scene-specific expertise of our BYE encoder, achieving a notable 7\% improvement and reaching a 95\% success rate in the simulator and 100\% in the real world in object association tasks.
    \item We evaluate our \ours{} in a photorealistic simulator AI2THOR~\cite{kolve2017ai2} as well as real-world environments and our method outperforms semantically expressive vision language foundation models. We release our code and dataset at \href{https://byencoder.github.io}{https://byencoder.github.io}.
\end{enumerate}

\section{Related Work}
\label{sec:related_work}

% Citations can be made using either \textbackslash citep\{\} or \textbackslash citet\{\}, depending from the appropriateness. To avoid the citation moving to the next line, it is often a good practice to replace the space before with a tilde (\~{}) character.
% Example 1: ``CoRL is the best conference ever~\citep{Gauss1857}.''
% Example 2: ``\citet{Lagrange1788} proved, both theoretically and numerically, that CoRL is the best conference ever.''

\textbf{Open-Vocabulary Semantic Mapping} In recent years, the advancement of vision language models and their fine-tuned counterparts~\cite{radford2021clip,liang2023ovseg,ghiasi2022openseg} have substantially innovated the mapping techniques in different robotic applications such as navigation~\cite{shah2023lm,huang23vlmaps,gu2024conceptgraphs,huang23avlmaps,werby2024hovsg}, manipulation~\cite{huang2023voxposer}, and 3D semantic scene understanding~\cite{kerr2023lerf,qin2024langsplat}. By integrating visual-language features into sparse topological nodes~\cite{shah2023lm,gu2024conceptgraphs,werby2024hovsg}, dense 3D voxels or 2D pixels~\cite{huang23vlmaps,OpenScene}, discrete 3D locations~\cite{conceptfusion}, or implicit neural representations~\cite{kerr2023lerf,engelmann2024opennerf,qin2024langsplat,kim2024garfield}, those created maps can be used to retrieve concepts with natural language descriptions, extending closed-set semantics retrieval~\cite{mccormac2018fusion++,xu2019mid} to open-vocabulary level and enabling more flexible and efficient human-robot interaction in the downstream tasks. However, most of the open-vocabulary semantic mapping approaches assumes a static environment, struggling to readjust the contents to scene changes in a long-term dynamic environment. In this work, we propose training a scene-wise point cloud encoder to extract class-agnostic, instance-level features stored in an object memory bank to manage future observations of dynamic scene changes.

% - LM-Nav
% - CoW
% - VLMaps
% - NLMap-SayCan
% - CLIP-Fields
% - ConceptFusion
% - OpenScene
% - OpenMask3D
% - OpenNERF
% - FM-Fusion
% - ConceptGraphs
% - HOV-SG
% - LERF
% - OpenNerf
% - LangSplat
% - GARField
% - 3D-LLM
% - FMGS

\textbf{Dynamic Scene Understanding} 

Understanding dynamic environments remains a challenge due to constant scene motion, complicating tracking and mapping. Early methods used semantic segmentation masks to filter objects during SLAM optimization~\cite{runz2017co,bescos2018dynaslam,yu2018ds}, but these rely on static/dynamic category assumptions, which are often invalid (e.g., parked vs. moving cars). Other approaches leveraged optical flow for object motion tracking~\cite{zhang2020flowfusion} or non-rigid tracking~\cite{newcombe2015dynamicfusion}. Recent advances in neural implicit representations, like Gaussian Splatting~\cite{kerbl2023gaussiansplatting}, enable rendering dynamic scenes over time~\cite{luiten2023dynamic} or simulating dynamics~\cite{xie2024physgaussian}, but focus on limited spatial ranges.

Long-term scene changes, central to this work, involve tasks like change detection and association, namely, tracking object displacement, addition, or removal across sequences. Prior methods registered and compared reconstructed scenes using visual, geometric, or semantic clues~\cite{adam2022objects,fu2022robust,adam2023has}, or trained networks for object association and reconstruction~\cite{zhu2024living,qiu2020indoor}. Others employed scene graphs~\cite{looper20233d} or probabilistic pipelines~\cite{Schmid-RSS24-Khronos}. These methods often relied on large synthetic datasets or predefined categories. In contrast, our approach trains an encoder on observational data sequences, avoiding category priors and synthetic data, to effectively tackle long-term change associations in dynamic scenes.

\textbf{Shape Representation Learning} A key technique to enhance the understanding of 3D scenes or objects is 3D shape representation learning. Early approaches focusing on 3D semantic understanding learned global or point-wise representations of point clouds~\cite{qi2017pointnet,wang2019dgcnn}, which nowadays serve as strong backbones for advanced methods. Other works learned implicit functions resembling traditional 3D shape representations like SDFs and occupancy grids with neural networks~\cite{park2019deepsdf}. Stemming from the techniques above, other methods explore learning the neural descriptor fields that map spatial locations relative to a shape to latent features for shape completion~\cite{chen2019learning}, registration~\cite{lin2023coarse}, manipulation~\cite{simeonov2022neural}, and object-level SLAM~\cite{fu2023neuse}. While the methods in those applications above have shown promising results, most of them require a curated 3D shape dataset containing a variety of complete shapes spanning different categories. In this work, our method only needs the observation data during the exploration of a scene to train a scene-wise encoder which can be used for object association in a long-term dynamic environment. By following the strategy introduced in SimCLR~\cite{chen2020simple}, we train an efficient encoder that attracts partial point cloud observations of the same object while repelling those of different instances.

\begin{figure*}[t]
\centering
\includegraphics[width=1\textwidth]{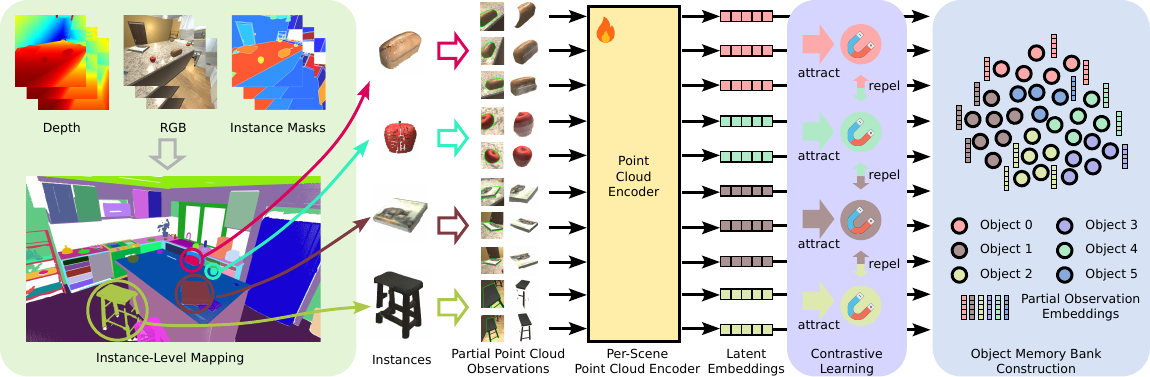}
\caption{Overview of the pipeline of \ours{} for long-term dynamic environment understanding. With the reference trial of exploration data, we first build an instance-level map using the RGB, depth, instance masks, and odometry inputs, from which we generate a partial object point cloud observations dataset. Later, we exploit the principles of contrastive learning to train a point cloud encoder from scratch. Finally, we encode all the partial observations in the dataset into latent embeddings and associate them with instance labels in the reference exploration trial as the object memory bank.}
\label{fig:overview}
\vspace{-1em}
\end{figure*}

\section{Problem Definition}
\label{sec:problem_definition}

To formalize the problem, we define a long-term dynamic environment where object locations can change between exploration trials, but each object remains static during a single trial. At each time step of a trial, we assume access to RGB images $\mathbf{I}_t$, depth images $\mathbf{D}_t$, 2D instance masks $\mathcal{M}_t=\{\mathbf{M}_{tk}\}_{k=1}^{K}$, and camera poses $\mathbf{T}_t$, from which we construct an instance-level map representing objects as independent point clouds $\{\mathcal{P}_i\}_{i=1}^{M}$, where $\mathcal{P}_i$ is the point cloud of the $i$-th object.

The problem of long-term object change detection and association is as follows: given two sets of point clouds, $\{\mathcal{P}^{ref}_i\}_{i=1}^{M}$ from a reference trial and $\{\mathcal{P}^{new}_j\}_{j=1}^{M'}$ from a new trial after object changes, find a bijective mapping $f: \{1, \dots, M'\} \to \{1, \dots, M\}$. If $f(j) = i$, $\mathcal{P}^{new}_j$ and $\mathcal{P}^{ref}_i$ correspond to the same object. Here, $M = M'$ after excluding added or removed objects.

\section{Method}
\label{sec:method}
This work aims to train a per-scene point cloud encoder that extracts latent features of partial object point cloud observations. With contrastive learning, we ensure that observations from the same object have similar embeddings while those from different objects are distinct from one another. By using the encoder to generate latent embeddings of all partial point cloud observations of all instances in the reference trial of the exploration, we can build a memory bank for the scene objects. Later, during the new trial of exploration after object relocations have happened, we can use the same encoder to get the latent embeddings of new point cloud observations, find nearest neighbors in the memory bank, and thus associate to an instance in the reference trial.

The idea of training a ``per-scene'' encoder is inspired by LangSplat~\cite{qin2024langsplat} that trained an autoencoder mapping high dimensional CLIP features~\cite{radford2021clip} for images collected in a single scene to low dimensional ones, which accelerates the optimization of CLIP-enriched Gaussian Splatting. In our work, we follow the idea of training a network only for data collected in a single scene during one trial of exploration which facilitates object change detection and association in long-term dynamic environments.

In the following, we introduce (i) the construction of an instance-level map which is the prerequisite of generating training data (Sec.~\ref{subsec:instance_level_map_construction}), (ii) the generation of partial point cloud observation dataset (Sec.~\ref{subsec:partial_pc_generation}), (iii) the training method of a per-scene object point cloud encoder with contrastive learning (Sec.~\ref{subsec:train_encoder}), (iv) the generation of memory bank (Sec.~\ref{subsec:object_memory_bank_creation}), and lastly (v) object association with the memory bank and the ensembling technique to further boost the object association performance (Sec.~\ref{subsec:object_association}). The overview of the pipeline is shown in Fig~\ref{fig:overview}.

\subsection{Instance-Level Map Construction}
\label{subsec:instance_level_map_construction}
We construct the instance-level map with the reference trial of exploration before object relocation. Given the instance segmentation masks $\mathcal{M}_t =\{\mathbf{M}_{tk}\}_{k=1,\ldots, K}$, depth image $\mathbf{D}_t$, and camera pose $\mathbf{T}_t$ of each frame $t$ in a trial of exploration, we can easily back-project the instance masks into 3D through the depth image, transform them to the global coordinate frame, and either fuse them with existing global instance point clouds with the instance IDs of those masks or initialize new global instances. After iterating the process in each frame, we can obtain a list of point clouds $\{\mathcal{P}_i\}_{i=1,\ldots, M}$ with instance IDs $i={1,\ldots, M}$ in this trial of exploration. For simplicity, we assume known instance masks and odometry as input to emphasize the effectiveness of our trained encoder and mitigate the impact of the quality of segmentation and odometry results. However, this can be easily extended with any existing instance-level mapping techniques such as ConceptGraphs~\cite{gu2024conceptgraphs}, HOV-SG~\cite{werby2024hovsg}, and so on.

\subsection{Partial Point Cloud Observation Data Generation}
\label{subsec:partial_pc_generation}
After obtaining the instance-level map $\{\mathcal{P}_i\}_{i=1,\ldots, M}$, for each instance point cloud $\mathcal{P}_i$, we find all the instance masks used to generate it and back-project them into camera coordinate frame to get a list of partial point cloud observations of the instance $\{\mathcal{P}^{cam}_{ir}\}_{r=1,\ldots, R_i}$ where $R_i$ is the total number of masks for object $i$ during this trial of exploration. Each $\mathcal{P}^{cam}_{ir}$ contains a list of 6-dimensional vectors each storing the 3D position and the RGB values of a point. We then subtract each point cloud's 3D coordinate with its mean to obtain a zero-center point cloud $\bar{\mathcal{P}}_{ir}$ where $\mathbf{x}_{\bar{\mathcal{P}}} = \mathbf{x}_{\mathcal{P}} - \frac{1}{|\mathcal{P}|} \sum_{\mathbf{x} \in \mathcal{P}} \mathbf{x}$ where $\mathbf{x}_{\bar{\mathcal{P}}} \in \mathbb{R}^3$ and $\mathbf{x}_{\mathcal{P}} \in \mathbb{R}^3$. For each zero-center point cloud $\bar{\mathcal{P}}_{ir}$, we take its instance id $i$ as label, forming one data sample as a tuple $(\bar{\mathcal{P}}_{ir}, i)$. For simplicity, we denote all point clouds and their object ID labels as $\{(\bar{\mathcal{P}}_{k}, y_k)\}_{k=1,\ldots, L}$ where $y_k$ is the object ID, and $L=\sum_{i=1}^{M} R_i$ is the total masks number of all objects.

\subsection{Training Scene-wise Object Point Cloud Encoder}
\label{subsec:train_encoder}

\begin{figure*}[t]
\centering
\includegraphics[width=1\textwidth]{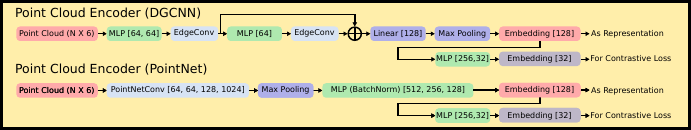}
\caption{The architecture of the point cloud encoder. We first follow the architecture of DGCNN~\cite{wang2019dgcnn} and PointNet~\cite{qi2017pointnet} with the training scheme of SimCLR~\cite{chen2020simple}, which add one more MLP layer without normalization following of the embedding output layer and project the representation to low dimensional space for more efficient contrastive learning.}
\label{fig:encoder}

\vspace{-2em}
\end{figure*}

The main goal of the encoder is to extract a latent representation for a point cloud in a scene so that the point clouds belonging to the same object have similar embeddings while the point clouds from different objects are far away from one another in the embedding space. In Sec.~\ref{subsec:partial_pc_generation}, we obtain a dataset of point cloud and object label pairs. In this section, we will walk through the pipeline of training the encoder with contrastive learning. Following the idea of SimCLR~\cite{chen2020simple}, the training pipeline is comprised of five major components: (i) a data preprocessing step, (ii) a stochastic data augmentation module, (iii) a neural network base encoder $\mathcal{E}(\cdot)$, (iv) a small neural network projection head $\mathbf{g}(\cdot)$, and (v) a contrastive loss function. The architecture of the encoder is shown in Fig.~\ref{fig:encoder}. In the following, we will walk through these components.

\textbf{Data Preprocessing:} Due to the variation in the size of different objects, we need to sample the point clouds to ensure the balance of the training workload. In this work, we use a mixed sampling strategy. For point clouds with more than 1024 points, we first apply voxel-downsampling to a resolution of 0.01 meter. If the points number still exceeds 1024, we apply farthest point sampling to select 1024 points. In this way, we guarantee that the points number is below 1024 and ensure an efficient training process.

\textbf{Data Augmentation:} In this work, we sequentially apply random jittering of the point positions with a range of 0 to 0.03 meter, rotation of the whole point cloud around the X-axis, around the Y-axis, and around the Z-axis of the point cloud $\bar{\mathcal{P}}_k$. The rotation around each axis has a range of 0 to 30 degrees.

\textbf{The Neural Network Base Encoder.} We choose the architectures of DGCNN~\cite{wang2019dgcnn} and PointNet~\cite{qi2017pointnet} as the backbones of our encoder because of DGCNN's potential to dynamically capture local geometric relationships and the simplicity of the PointNet. Furthermore, the pipeline is compatible with other base encoder backbones~\cite{qi2017pointnet++}, allowing for potential improvements when better point cloud processing architectures are discovered. Both architectures are shown in Fig.~\ref{fig:encoder}. 

\textbf{Embedding projection head:} As in SimCLR~\cite{chen2020simple}, we project the embeddings $\mathbf{h}(\bar{\mathcal{P}})$ generated by the backbone above into a low dimensional space $\mathbf{g}(\bar{\mathcal{P}}) = MLP(\mathbf{h}(\bar{\mathcal{P}}))$ with an MLP layer for the ease of contrastive loss computation. During inference, we don't apply the final MLP projection head and directly use $\mathbf{h}(\bar{\mathcal{P}})$ as the point cloud representation.

\textbf{Contrastive Learning Loss:} In this work, we use the NT-Xent loss proposed in~\cite{sohn2016improved}: 

\begin{equation}
    \mathbb{L}_{i, j}=-\log \frac{\exp (\mathbf{g}_i, \mathbf{g}_{+}) / \tau}{\sum_{k=1}^{K} \exp (\mathbf{g}_i, \mathbf{g}_k / \tau)}
\end{equation}

where $\mathbf{g}_i$ is the anchor embedding, $\mathbf{g}_{+}$ is the positive sample embedding which has the same object ID as $\mathbf{g}_i$ while all other $\{\mathbf{g}_k\}_{k=1,\ldots, K, k\neq i}$ are negative samples in the batch that have different object IDs from $\mathbf{g}_i$.

\textbf{Training Details.} During each training iteration, we randomly load a batch of 64 data samples, along with one additional positive sample for each instance (i.e., a sample with the same object ID), resulting in 128 samples in total and ensuring at least 64 positive pairs. For each instance, all other instances in the batch, except its positive counterpart, are treated as negative samples. The model is trained with a learning rate of 0.003 for 300 epochs, using a 90/10 training-validation split. We evaluate the validation loss every 300 iterations and save the checkpoint with the lowest validation loss for use in experiments. In the dynamic edge convolutional network, we set $k=10$ for the k-NN search.

\subsection{Object Memory Bank Generation}
\label{subsec:object_memory_bank_creation}

After training the encoder for the scene with the reference trial of exploration data, you can encode all the partial point cloud observations $\{\bar{\mathcal{P}}_{l}\}_{l=1,\ldots, L}$ in your dataset in Sec.~\ref{subsec:partial_pc_generation} into the latent embeddings $\{\mathbf{h}(\bar{\mathcal{P}}_{l})\}_{l=1,\ldots, L}$ and associate those embeddings with their corresponding instance ID labels $\{y_l\}_{l=1,\ldots, L}$, forming embedding-ID pairs $\{(\mathbf{h}^{ref}_{l}, y^{ref}_l)\}_{l=1,\ldots, L}$ (for simplicity, we write $\mathbf{h}(\bar{\mathcal{P}}_{l})$ as $\mathbf{h}^{ref}_{l}$). Since the embeddings and labels are for the reference exploration trial, we add a superscript of $ref$ to their symbols. We treat these embeddings and labels as the object memory bank of the scene. Since the object embeddings are translation-invariant, we can easily look up the memory bank when new observations come after object location changes. 

\subsection{Object Association in Dynamic Environment}
\label{subsec:object_association}

\textbf{Retrieval from Object Memory Bank} Now we switch to the new trial of exploration data, after a random number of object locations changes without adding new objects or removing old ones. Given the instance segmentation masks $\mathcal{M}_t =\{\mathbf{M}_{tk}\}_{k=1,\ldots, K}$, depth image $\mathbf{D}_t$, and camera pose $\mathbf{T}_t$ of each frame $t$ in a new trial of exploration, we can build a new instance-level map as in Sec.~\ref{subsec:instance_level_map_construction}. In addition, for each mask observation $\mathbf{M}_{tk}$, we can back-project them into the camera coordinate, center them at their means, and apply voxel-downsampling to a resolution of 0.01 meter to obtain partial point cloud observations $\bar{\mathcal{P}}^{new}_{tk}$ as in Sec.~\ref{subsec:partial_pc_generation}. We can use the trained encoder (see Sec.~\ref{subsec:train_encoder}) to generate a latent embedding $\mathbf{h}(\bar{\mathcal{P}}^{new}_{tk})$ (we write it as $\mathbf{h}^{new}_{tk}$ for simplicity) for each partial point cloud observation $\bar{\mathcal{P}}^{new}_{tk}$. In the following step, we can find the distance of $\mathbf{h}^{new}_{tk}$ to all the embeddings $\{\mathbf{h}^{ref}_{l}\}_{l=1,\ldots, L}$ in the object memory bank created in Sec.~\ref{subsec:object_memory_bank_creation}. We take the 10 nearest embeddings and store their object ID labels in the reference trial. These reference labels are associated with the global object ID in the new trial in a dictionary $\{new\ object\ ID:\ reference\ object\ IDs\}$. Whenever there are new observations for the same new object ID later, the reference object IDs list will be extended with new labels. During this process, we can use frequency to approximate the probability of association such that $P(f(j)=i|\mathbf{z}_{t=1:n}) = \frac{\#(y^{ref} = i)}{\sum^{M}_{m=1} \#(y^{ref} = m)}$ where $f(\cdot)$ is the mapping from new trial object ID $j$ to reference trial object ID $i$, $\mathbf{z}_t$ denotes the observations at timestep $t$, $\#(y^{ref} = i)$ represents the count of reference labels $i$ associated with new object $j$, and $M$ denotes the total number of objects in reference trial. To determine the final association for a new object ID, we simply retrieve the reference object ID with the highest probability.

\begin{figure}[t]
\centering
\includegraphics[width=0.49\textwidth]{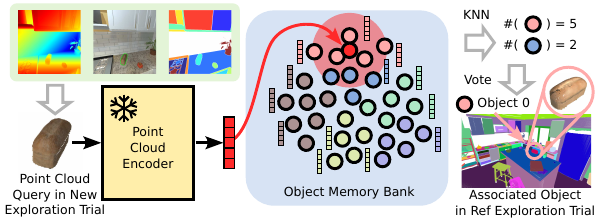}
\caption{The process of querying the object memory bank with new exploration trial data. Given the RGB, depth, and instance masks in the new exploration trial, we extract the partial point cloud observation, encode the point cloud with the pre-trained per-scene point cloud encoder as in Sec.~\ref{subsec:train_encoder}, and obtain a latent embedding which we use to look up the object memory bank (see Sec.~\ref{subsec:object_memory_bank_creation}) and find the K nearest neighboring embeddings. After counting the neighboring embeddings' instance labels, we can associate the partial observation to an instance in the reference trial of exploration.}
\label{fig:query}
\vspace{-1em}
\end{figure}

\textbf{VLM Ensembling} To further improve the object association success rate, we propose an ensembling scheme to benefit from both the semantic reasoning capability of Vision Language Models as well as the scene-specific expertise of \ours{} encoder. With the association dictionary $\{new\ object\ ID:\ reference\ object\ IDs\}$, we can construct a matrix storing normalized scores between each new object and reference object $A^{BYE}$ where $a^{BYE}_{ji}$ indicates the score of new object $j$ associating with reference object $i$, which is $P(f(j)=i|\mathbf{z}_{t=1:n})$ introduced above. To exploit the power of VLMs, we can use HOV-SG~\cite{werby2024hovsg} to construct instance-level open-vocabulary maps for both the reference and the modified scenes. In general, HOV-SG first constructs a voxel map where each voxel stores the back-projected open-vocabulary visual features generated by Vision Language Models such as CLIP~\cite{radford2021clip}. Then it creates a segment-level geometric map where each object is represented by an independent point cloud. To obtain an open-vocabulary feature for each object, we simply collect the nearest voxel features in the voxel map for all points in each object's point cloud and apply DBSCAN to select the major feature cluster. Finally, we take the feature closest to the cluster mean as the object's feature. By computing the cosine similarity between the features of new and old objects, we can also obtain a score matrix $A^{VLM}$. We ensemble their results by taking the sum of the two matrices $A=A^{BYE}+A^{VLM}$ and applying the Hungarian algorithm to get the association results. 

\section{Experimental Results}
\label{sec:result}
In this section, we want to answer the following three questions: (i) how powerful \ours{} is compared to the foundation-model-based encoder in object association tasks in long-term dynamic environments (Sec.~\ref{exp:object_association}), (ii) how the object association ability of \ours{} generalizes to real-world environments across various scenes (Sec.~\ref{exp:real_world_association}), and (iii) how efficient \ours{} is for potential robotics applications (Sec.~\ref{exp:runtime_analysis}). We will address all these questions in the following subsections.
% The goal of our experiments are four-fold: (i) demonstrate how effective the encoder is in object association tasks in long-term dynamic environments (Sec.~\ref{exp:object_association}), (ii) show how our encoder enables 3D shape prompt for object retrieval (Sec.~\ref{exp:3d_shape_prompt}), (iii) justify our design choice through ablative study of different encoder architectures, and (iv) show the efficiency of our method (Sec.~\ref{exp:runtime_analysis}).

\subsection{Object Association in Long-Term Dynamic Environments}
\label{exp:object_association}

\textbf{Experiment Setup:} To evaluate the association effectiveness of \ours{}, we collected 10 reference scenes (FloorPlan 1, 2, 4, 5, 6, which are kitchens, and FloorPlan 301-305, which are bedrooms) and 10 corresponding change scenes in AI2THOR~\cite{kolve2017ai2} where only object locations are changed in the same scene. A robot was manually controlled to gather exploration data, including RGB images, depth images, instance masks, and camera poses. For the change scenes, we initialized them as reference scenes, randomly moved some objects, and then collected exploration data after the changes. For each reference scene, we built an instance-level map, generated a dataset of partial instance point clouds with labels, trained the encoder, and used the checkpoint with the best validation loss to create an object memory bank as in Sec.~\ref{sec:method}. During the association process, we iterate through all observations of the new exploration, extracting masked point clouds, back-projecting, centering them, and generating latent embeddings using the same checkpoint. We then found the 10 nearest neighbors in the object memory bank for each new observation, maintaining a count for each associated reference instance ID. This created a mapping from new instance IDs to reference IDs which evolves over time when more observations come. Finally, for pure \ours{} methods, namely \ours{} (DGCNN) and \ours{} (PointNet), we used majority voting to assign the most likely reference instance ID to each new instance. For ensemble methods \ours{} (DGCNN+CLIP) and \ours{} (PointNet+CLIP), we follow the VLM ensembling trick introduced in Sec.~\ref{subsec:object_association} and the Hungarian algorithm to assign a reference instance ID to each object in the new exploration.

\begin{table*}[thb]
\centering
\scriptsize
\setlength\tabcolsep{2.5pt}
\caption{\textsc{Object Association Success Rate in AI2THOR Long-Term Dynamic Environments}}
\begin{threeparttable}
    \begin{tabular}{lc|ccccccccccccccc}
     \toprule
     
    \multirow{3}[1]{*}{Method} & \multicolumn{16}{c}{Success Rate (\%)} \\
    \cmidrule(lr){2-17}
        &   Overall          & Garbage Can        &  Bowl & Mug &  Cell Phone  & Credit Card  & Stool   & Bread & Desk  & Key Chain  & Alarm Clock  & Laptop  & Pan & Pen & Salt Shaker & Spoon \\
        & (252) & (10) & (9) & (9) & (6) & (6) & (6) & (5)  & (5) & (5) & (5) & (5) & (5) & (5) & (5) & (4)\\

    \midrule
 
    DINOv2~\cite{oquab2023dinov2} & 50.3 & 70 & 55.6 & 33.3 & 16.7 & 50 & 83.3  & 20 & 80 & 60 & 20 & 60 & 40 & 40 & 20 & 75\\
    
    LSeg~\cite{li2022lseg}        & 50.0 & 80 & 55.6 & 44.4 & 16.7 & 0 & \textbf{100} & 40 & 40 & 40 & 40 & \textbf{100} & 20 & 0 & 40 & 25\\
    
    OVSeg~\cite{liang2023ovseg}   & 75.4 & \textbf{100} & 77.8 & 77.8 & 50 & 50 & \textbf{100}  & 60 & 80 & 60 & 60 & \textbf{100} & 40 & 60 & 40 & 75\\
    
    CLIP~\cite{radford2021clip}   & 88.9 & \textbf{100} & \textbf{100} & 88.9 & 83.3 & \textbf{100} & \textbf{100} & \textbf{100} & \textbf{100} & \textbf{100} & \textbf{80} & \textbf{100} & 80 & 20 & 40 & 50 \\ 
    
    \midrule
    \ours{} (DGCNN)               & 82.9 & 90 & 66.7 & 77.8 & \textbf{100} & \textbf{100} & \textbf{100} & \textbf{100} & \textbf{100} & \textbf{100} & \textbf{80} & 60 & \textbf{100} & 0 & 80 & 75\\
    
    \ours{} (PointNet)            & 85.7 & 90 & 77.8 & 77.8 & 66.7 & \textbf{100} & \textbf{100} & \textbf{100} & 80 & \textbf{100} & \textbf{80} & 60 & \textbf{100} & 40 & 80 & \textbf{100}\\
    
    \ours{} (DGCNN + CLIP)        & 92.5 & \textbf{100} & 88.9 & 77.8 & \textbf{100} & \textbf{100} & \textbf{100} & \textbf{100} & \textbf{100} & \textbf{100} & \textbf{80} & \textbf{100} & \textbf{100} & 60 & \textbf{100} & \textbf{100} \\
    
    \ours{} (PointNet + CLIP)     & \textbf{95.6} & \textbf{100} & 88.9 & \textbf{100} & \textbf{100} & \textbf{100} & \textbf{100} & \textbf{100} & \textbf{100} & \textbf{100} & \textbf{80} & \textbf{100} & \textbf{100} & \textbf{80} & \textbf{100} & \textbf{100} \\
    
    \bottomrule
    \end{tabular}
    The overall and per-class object association success rates in 10 AI2THOR scenes. The number in the parenthesis denotes the total occurrence number of a specific kind of object.
\end{threeparttable}
\label{tab:association}
\vspace{-2em}
\end{table*}

\textbf{Baselines:} The goal of the baseline methods is to integrate semantic-rich visual language features into the segment-level map, namely, associate one feature with each instance. We exploit the open-vocabulary mapping scheme introduced in HOV-SG~\cite{werby2024hovsg} as is introduced in the VLM ensembling part in Sec.~\ref{subsec:object_association}. First, we build a voxel feature map. This requires a visual encoder that can generate dense pixel-level visual features. Here we use LSeg~\cite{li2022lseg}, OVSeg~\cite{liang2023ovseg}, DINOv2~\cite{oquab2023dinov2} (ViT B/14), and CLIP~\cite{radford2021clip} (ViT B/32) as the encoders in our baselines. We use DINOv2's last intermediate layer's patch-wise embeddings and upsample it to the image resolution with nearest-neighbor interpolation. CLIP originally only generates global image features but we can obtain small regions of a single image based on the instance masks and assign each region's CLIP embedding to all pixels in its mask to obtain dense visual language features. Then we can back-project depth pixels into 3D space, find the voxel they belong to, and integrate the pixel features into the voxel by running average fusion. Second, we construct an instance-level map with the instance masks. Finally, after the voxel map and the instance map are completed, we need to assign one visual language feature to each instance. We first search for the nearest neighboring voxel for each point in an instance, then collect the voxel's associated feature. After collecting all features for all points in one instance, we apply DBSCAN~\cite{ester1996density} to the features and find the major cluster's mean. Then we find the feature with the closest distance to the cluster mean as the instance's feature. We build such instance-level feature maps for reference and new exploration trials for each scene. During association, we simply collect all instance features in the reference and the new scenes and compute the cosine similarity between each pair of objects, forming a score matrix $A^{VLM}$. Finally, we use the Hungarian algorithm based on $A^{VLM}$ to determine the association.

\textbf{Metrics:} We use the association success rate as our metric, which can be defined as the number of correctly associated objects divided by the total number of objects. In this experiment, we want to test the capability of associating movable objects whose total number is 252, including 63 object categories. We further listed partial per-class success rates for 15 common categories with various sizes and shapes.

\textbf{Results:} The results are shown in Table~\ref{tab:association}. The first two columns show the methods and the overall success rates. The rest of the columns show the per-class success rates for movable objects with high frequencies of occurrence. We observe that the PointNet version of \ours{}, ensembling with CLIP, outperforms all other baselines by a large margin with roughly 7\% improvement compared to pure foundation-model-based methods, achieving a 95.6\% object association success rate.

We further analyzed why ensembling \ours{} performs better than other foundation-model-based embeddings by looking at per-class association results. For objects that are commonly occurring in various datasets like ``Garbage Can'', ``Bowl'', ``Stool'', ``Bread'', ``Desk'', and ``Laptop'', foundation models perform very well in the association tasks, sometimes even better than \ours{}. The major reason behind this might be that the pre-training and fine-tuning processes allow the model to learn reliable and robust features for those categories. Another reason might be that these objects are large and their features are salient even without background context. However, when we look at long-tailed categories or relatively small objects like ``Cell Phone'', ``Pen'', ``Spoon'', and ``Salt Shaker'', foundation models struggle to generate reliable features without clear backgrounds or zoom-in pictures, and thus fail to correctly associate them. In contrast, our ensembling \ours{} takes advantage of both the foundation models' general semantic understanding as well as the expertise scene-related features learned from the customized data, consistently outperforming all baselines. We further show the benefits of such combinations in real-world experiments in the next subsection (Sec.~\ref{exp:real_world_association}).

% \subsection{3D Shape Prompt Object Retrieval}
% \label{exp:3d_shape_prompt}

% \subsection{Spatio-Temporal Open-Vocabulary Navigation}
% \label{exp:spatio_temporal_navigation}

% \subsection{Real World Object Association in Long-Term Dynamic Environments}
% \label{exp:real_world_association}

% \subsection{Architecture Ablation}
% \label{exp:ablation}
\begin{figure}[!th]
\centering
\includegraphics[width=0.49\textwidth]{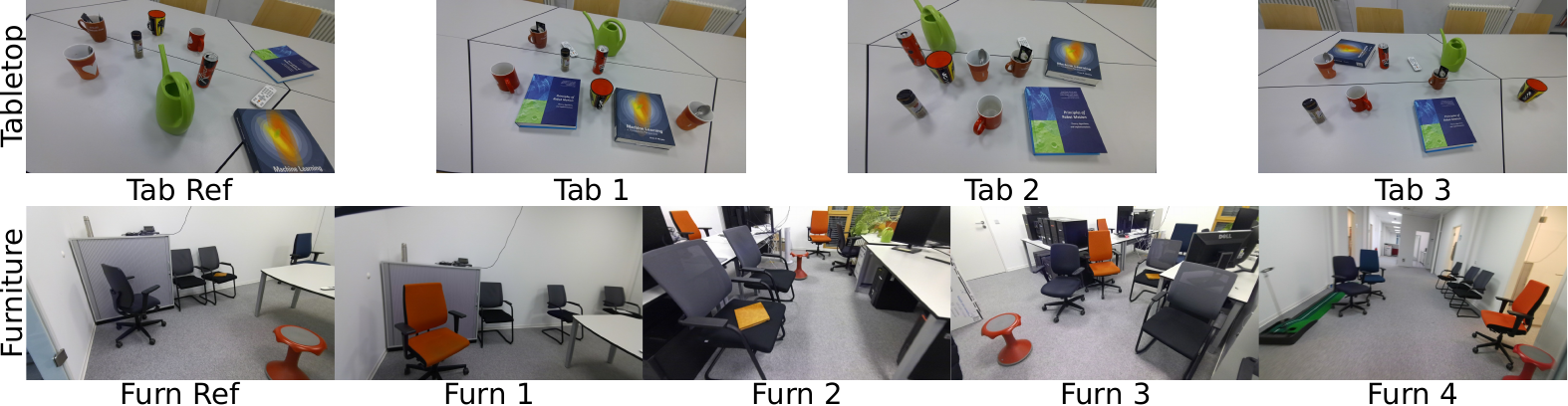}
\caption{The tabletop and furniture setups in the real world. We take one sequence of data in each setup as the reference and the rest as the test data.}
\label{fig:real_world_setup}
\vspace{-1em}
\end{figure}

\subsection{Real-World Experiments}
\label{exp:real_world_association}

\textbf{Experiment Setup:} We evaluated \ours{}'s object association performance in two real-world scenarios: a tabletop setting and a furniture setting, reflecting manipulation and navigation tasks. In the tabletop setting, 10 objects were arranged in four random layouts on a table. In the furniture setting, eight chairs were positioned across three rooms (meeting room, office, hallway) in five layouts, with variations such as objects placed on seat pads (e.g., paper, notebook, or empty). This setup, shown in Fig.~\ref{fig:real_world_setup}, is particularly challenging due to the chairs' similar appearances. Using an Azure Kinect sensor, we recorded nine exploration trials, generating instance masks with SAM2~\cite{ravi2024sam2} and estimating camera poses via DROID-SLAM~\cite{teed2021droid}. One trial per setting was used to train the scene-specific encoder, with the rest for evaluation. The total association task number is 62.

\begin{figure}[!th]
\centering
\includegraphics[width=0.49\textwidth]{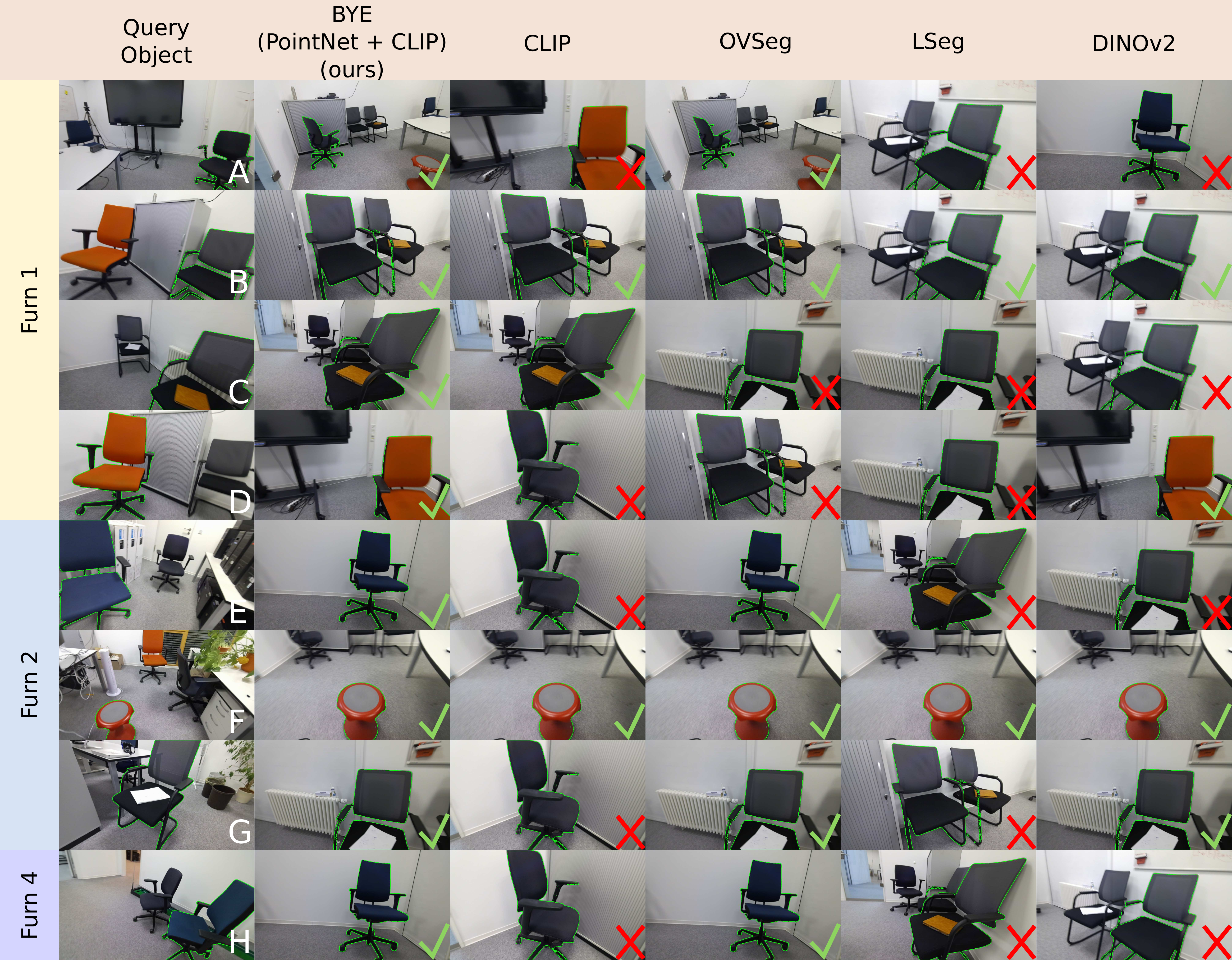}
\caption{The qualitative object association results in furniture setup in the real world. In the first column, we visualize the query objects in new exploration trials and their masks (green contours). In the following columns, we visualize the associated objects in the reference exploration trial and their masks. We compare \ours{} (PointNet + CLIP) ensemble method with baselines using CLIP, OVSeg, LSeg, and DINOv2 features from vision foundation models.}
\label{fig:real_world_result}

\end{figure}

\textbf{Results:} The results are shown in Table~\ref{tab:real_world_association}. In tabletop settings where there are no duplicate objects from the same categories, both CLIP and \ours{} perform perfectly. Fine-tuned foundation models such as LSeg and OVSeg struggle in these objects which are long-tailed in the datasets they are fine-tuned on. In the furniture setting, we can see a clearer benefit of combining \ours{} and foundation features. When CLIP struggles to differentiate textures and attributes of the same class of objects and pure \ours{} fails to capture all common sense semantic features, the ensembling \ours{} achieves perfect association by combining both. We further show the qualitative results in Fig.~\ref{fig:real_world_result}. 

\begin{table}[thb]
\centering
\scriptsize
\setlength\tabcolsep{2.5pt}
\caption{\textsc{Object Association Success Rate in Real-World Long-Term Dynamic Environments}}
\begin{threeparttable}
    \begin{tabular}{lccccccc}
     \toprule
     
    \multirow{3}[1]{*}{Method} & \multicolumn{7}{c}{Success Rate (\%)} \\
    \cmidrule(lr){2-8}
        &   Tab 1 & Tab 2 & Tab 3 & Furn 1 & Furn 2 & Furn 3 & Furn 4\\

    \midrule
 
    DINOv2~\cite{oquab2023dinov2} & 40 & 40 & 40 & 37.5 & 25 & 37.5 & 50\\
    
    LSeg~\cite{li2022lseg}        & 40 & 40 & 40 & 25 & 12.5 & 25 & 25\\
    
    OVSeg~\cite{liang2023ovseg}   & 80 & 80 & 80 & 62.5 & \textbf{100} & \textbf{100} & 75\\
    
    CLIP~\cite{radford2021clip}   & \textbf{100} & \textbf{100} & \textbf{100} & 62.5 & 62.5 & \textbf{100} & 62.5\\ 
    
    \midrule
    \ours{} (DGCNN)               & \textbf{100} & \textbf{100} & \textbf{100} & \textbf{100} & 87.5 & 87.5 & 87.5 \\
    
    \ours{} (PointNet)            & \textbf{100} & \textbf{100} & \textbf{100} & 75 & \textbf{100} & \textbf{100} & 62.5 \\

    \ours{} (DGCNN + CLIP)               & \textbf{100} & \textbf{100} & \textbf{100} & \textbf{100} & \textbf{100} & \textbf{100} & \textbf{100} \\

    \ours{} (PointNet + CLIP)               & \textbf{100} & \textbf{100} & \textbf{100} & \textbf{100} & \textbf{100} & \textbf{100} & \textbf{100} \\
    
    % \ours{} (DGCNN + CLIP)        & \\
    
    % \ours{} (PointNet + CLIP)     &  \\
    \bottomrule
    \end{tabular}
\end{threeparttable}
\label{tab:real_world_association}
\vspace{-2em}
\end{table}

\subsection{Runtime Analysis}
\label{exp:runtime_analysis}

\textbf{Experiment Setup:} We use a machine with an AMD Ryzen 9 PRO 7945 12-core CPU and an NVIDIA GeForce RTX 4060Ti GPU with 16 GB VRAM. We load the DGCNN and PointNet encoders in evaluation mode, and load the data with different batch sizes and a single CPU thread. Then we iterate through a dataset of each scene (41267 partial point cloud samples in total) and let the encoder generate latent embeddings in inference mode. We count the time for preprocessing (downsampling) and generating all embeddings with partial point cloud observations in the scene and divide the time by the total partial observations number to get the average runtime of the encoder.

\textbf{Results:} Table~\ref{tab:runtime} shows that the average runtime of the encoder considering the batch processing is around 11ms which is highly efficient to run at more than 85 samples per second, making it very low-cost to be plugged in any robotic pipelines without adding heavy burdens.

\begin{table}[h]
\centering
\scriptsize
\setlength\tabcolsep{2.5pt}
\caption{\textsc{Runtime Analysis of \ours{}}}
% \begin{threeparttable}
    \begin{tabular}{lcccc}
     \toprule
     
    \multirow{2}{*}{Method} & Batch & Total  & Average Runtime & Frequency \\
     & Size & Runtime (s) &  per Sample (ms) & (Samples/Sec) \\
    
    \midrule
    \ours{} (DGCNN)   & \multirow{2}{*}{1} & 690.4 & 16.7 & 59.8\\
    \ours{} (PointNet) &  & 805.1 & 19.7 & 50.7\\
    
    \midrule
    
    % \ours{} (DGCNN)    & \multirow{2}{*}{16} & 501.3 & 12.1 & 82.6\\
    % \ours{} (PointNet) &  & 496.8 & 12.0 & 83.3\\

    % \midrule
    
    \ours{} (DGCNN)    & \multirow{2}{*}{32} & 466.7 & 11.3 & 88.5\\
    \ours{} (PointNet) &  & 475.2 & 11.5 & 87.0 \\

    \bottomrule
    \end{tabular}
    % Higher values are better. \rebuttal{The used evaluation metrics are defined in Sec.~\ref{sec:evaluation-metrics-3d-seg-suppl} in the supplementary material.} The ConceptFusion pipeline evaluated against made use of instance masks predicted by SAM~\cite{kirillov2023segany}. \color{black} The MinkowskiNet~\cite{choy20194d} is a privileged method that was trained on the full set of ScanNet~\cite{dai2017scannet} scenes to demonstrate the gap between zero-shot and fully-supervised methods.
% \end{threeparttable}
\label{tab:runtime}
\vspace{-2em}
\end{table}

\section{Conclusion}
\label{sec:conclusion}
In this work, we presented a novel approach called \ours{} to the problem of object association in long-term dynamic scenes. By combining a per-scene encoder trained with scene-specific exploration data with foundation models, we achieve a near-perfect association success rate (95\% in simulation and 100\% in real-world scenarios) across dynamic scenes with high efficiency. We believe that our work highlights a promising pathway toward lifelong learning, namely customizing foundation models by combining them with a lightweight scene-specific expert model. However, \ours{} faces limitations, including challenges in detecting newly added or removed objects in long-term dynamic environments and reliance on instance masks as supervision for training the per-scene encoder. Future research will focus on integrating this encoder with instance-level open-vocabulary mapping techniques and deploying it in real-world robotic navigation systems. This integration has the potential to advance spatio-temporal open-vocabulary navigation and object search capabilities. Additionally, extending the proposed approach to training scene-specific models using exploration data to other application domains presents a promising avenue for further investigation. 

\vspace{-1em}
\bibliographystyle{IEEEtran}
\bibliography{reference}

\begin{thebibliography}{10}
\providecommand{\url}[1]{#1}
\csname url@samestyle\endcsname
\providecommand{\newblock}{\relax}
\providecommand{\bibinfo}[2]{#2}
\providecommand{\BIBentrySTDinterwordspacing}{\spaceskip=0pt\relax}
\providecommand{\BIBentryALTinterwordstretchfactor}{4}
\providecommand{\BIBentryALTinterwordspacing}{\spaceskip=\fontdimen2\font plus
\BIBentryALTinterwordstretchfactor\fontdimen3\font minus
  \fontdimen4\font\relax}
\providecommand{\BIBforeignlanguage}[2]{{%
\expandafter\ifx\csname l@#1\endcsname\relax
\typeout{** WARNING: IEEEtran.bst: No hyphenation pattern has been}%
\typeout{** loaded for the language `#1'. Using the pattern for}%
\typeout{** the default language instead.}%
\else
\language=\csname l@#1\endcsname
\fi
#2}}
\providecommand{\BIBdecl}{\relax}
\BIBdecl

\bibitem{o2024open}
A.~O’Neill, A.~Rehman, A.~Maddukuri, A.~Gupta, A.~Padalkar, A.~Lee,
  A.~Pooley, A.~Gupta, A.~Mandlekar, A.~Jain \emph{et~al.}, ``Open
  x-embodiment: Robotic learning datasets and rt-x models: Open x-embodiment
  collaboration 0,'' in \emph{2024 IEEE International Conference on Robotics
  and Automation (ICRA)}.\hskip 1em plus 0.5em minus 0.4em\relax IEEE, 2024,
  pp. 6892--6903.

\bibitem{runz2017co}
M.~R{\"u}nz and L.~Agapito, ``Co-fusion: Real-time segmentation, tracking and
  fusion of multiple objects,'' in \emph{2017 IEEE International Conference on
  Robotics and Automation (ICRA)}.\hskip 1em plus 0.5em minus 0.4em\relax IEEE,
  2017, pp. 4471--4478.

\bibitem{bescos2018dynaslam}
B.~Bescos, J.~M. F{\'a}cil, J.~Civera, and J.~Neira, ``Dynaslam: Tracking,
  mapping, and inpainting in dynamic scenes,'' \emph{IEEE Robotics and
  Automation Letters}, vol.~3, no.~4, pp. 4076--4083, 2018.

\bibitem{yu2018ds}
C.~Yu, Z.~Liu, X.-J. Liu, F.~Xie, Y.~Yang, Q.~Wei, and Q.~Fei, ``Ds-slam: A
  semantic visual slam towards dynamic environments,'' in \emph{2018 IEEE/RSJ
  international conference on intelligent robots and systems (IROS)}.\hskip 1em
  plus 0.5em minus 0.4em\relax IEEE, 2018, pp. 1168--1174.

\bibitem{xu2019mid}
B.~Xu, W.~Li, D.~Tzoumanikas, M.~Bloesch, A.~Davison, and S.~Leutenegger,
  ``Mid-fusion: Octree-based object-level multi-instance dynamic slam,'' in
  \emph{2019 International Conference on Robotics and Automation (ICRA)}.\hskip
  1em plus 0.5em minus 0.4em\relax IEEE, 2019, pp. 5231--5237.

\bibitem{adam2022objects}
A.~Adam, T.~Sattler, K.~Karantzalos, and T.~Pajdla, ``Objects can move: 3d
  change detection by geometric transformation consistency,'' in \emph{European
  Conference on Computer Vision}.\hskip 1em plus 0.5em minus 0.4em\relax
  Springer, 2022, pp. 108--124.

\bibitem{adam2023has}
A.~Adam, K.~Karantzalos, L.~Grammatikopoulos, and T.~Sattler, ``Has anything
  changed? 3d change detection by 2d segmentation masks,'' \emph{arXiv preprint
  arXiv:2312.01148}, 2023.

\bibitem{zhu2024living}
L.~Zhu, S.~Huang, K.~Schindler, and I.~Armeni, ``Living scenes: Multi-object
  relocalization and reconstruction in changing 3d environments,'' in
  \emph{Proceedings of the IEEE/CVF Conference on Computer Vision and Pattern
  Recognition}, 2024, pp. 28\,014--28\,024.

\bibitem{chang2015shapenet}
A.~X. Chang, T.~Funkhouser, L.~Guibas, P.~Hanrahan, Q.~Huang, Z.~Li,
  S.~Savarese, M.~Savva, S.~Song, H.~Su \emph{et~al.}, ``Shapenet: An
  information-rich 3d model repository,'' \emph{arXiv preprint
  arXiv:1512.03012}, 2015.

\bibitem{qin2024langsplat}
M.~Qin, W.~Li, J.~Zhou, H.~Wang, and H.~Pfister, ``Langsplat: 3d language
  gaussian splatting,'' in \emph{Proceedings of the IEEE/CVF Conference on
  Computer Vision and Pattern Recognition}, 2024, pp. 20\,051--20\,060.

\bibitem{kerbl2023gaussiansplatting}
B.~Kerbl, G.~Kopanas, T.~Leimk{\"u}hler, and G.~Drettakis, ``3d gaussian
  splatting for real-time radiance field rendering.'' \emph{ACM Trans. Graph.},
  vol.~42, no.~4, pp. 139--1, 2023.

\bibitem{kolve2017ai2}
E.~Kolve, R.~Mottaghi, W.~Han, E.~VanderBilt, L.~Weihs, A.~Herrasti, M.~Deitke,
  K.~Ehsani, D.~Gordon, Y.~Zhu \emph{et~al.}, ``Ai2-thor: An interactive 3d
  environment for visual ai,'' \emph{arXiv preprint arXiv:1712.05474}, 2017.

\bibitem{radford2021clip}
A.~Radford, J.~W. Kim, C.~Hallacy, A.~Ramesh, G.~Goh, S.~Agarwal, G.~Sastry,
  A.~Askell, P.~Mishkin, J.~Clark \emph{et~al.}, ``Learning transferable visual
  models from natural language supervision,'' in \emph{International conference
  on machine learning}.\hskip 1em plus 0.5em minus 0.4em\relax PMLR, 2021, pp.
  8748--8763.

\bibitem{liang2023ovseg}
F.~Liang, B.~Wu, X.~Dai, K.~Li, Y.~Zhao, H.~Zhang, P.~Zhang, P.~Vajda, and
  D.~Marculescu, ``Open-vocabulary semantic segmentation with mask-adapted
  clip,'' in \emph{Proceedings of the IEEE/CVF Conference on Computer Vision
  and Pattern Recognition}, 2023, pp. 7061--7070.

\bibitem{ghiasi2022openseg}
G.~Ghiasi, X.~Gu, Y.~Cui, and T.-Y. Lin, ``Scaling open-vocabulary image
  segmentation with image-level labels,'' in \emph{European Conference on
  Computer Vision}.\hskip 1em plus 0.5em minus 0.4em\relax Springer, 2022, pp.
  540--557.

\bibitem{shah2023lm}
D.~Shah, B.~Osi{\'n}ski, S.~Levine \emph{et~al.}, ``Lm-nav: Robotic navigation
  with large pre-trained models of language, vision, and action,'' in
  \emph{Conference on Robot Learning}.\hskip 1em plus 0.5em minus 0.4em\relax
  PMLR, 2023, pp. 492--504.

\bibitem{huang23vlmaps}
C.~Huang, O.~Mees, A.~Zeng, and W.~Burgard, ``Visual language maps for robot
  navigation,'' in \emph{Proceedings of the IEEE International Conference on
  Robotics and Automation (ICRA)}, London, UK, 2023.

\bibitem{gu2024conceptgraphs}
Q.~Gu, A.~Kuwajerwala, S.~Morin, K.~M. Jatavallabhula, B.~Sen, A.~Agarwal,
  C.~Rivera, W.~Paul, K.~Ellis, R.~Chellappa \emph{et~al.}, ``Conceptgraphs:
  Open-vocabulary 3d scene graphs for perception and planning,'' in \emph{2024
  IEEE International Conference on Robotics and Automation (ICRA)}.\hskip 1em
  plus 0.5em minus 0.4em\relax IEEE, 2024, pp. 5021--5028.

\bibitem{huang23avlmaps}
C.~Huang, O.~Mees, A.~Zeng, and W.~Burgard, ``Audio visual language maps for
  robot navigation,'' in \emph{Proceedings of the International Symposium on
  Experimental Robotics (ISER)}, Chiang Mai, Thailand, 2023.

\bibitem{werby2024hovsg}
A.~Werby, C.~Huang, M.~Büchner, A.~Valada, and W.~Burgard, ``{Hierarchical
  Open-Vocabulary 3D Scene Graphs for Language-Grounded Robot Navigation},'' in
  \emph{Proceedings of Robotics: Science and Systems}, Delft, Netherlands, July
  2024.

\bibitem{huang2023voxposer}
W.~Huang, C.~Wang, R.~Zhang, Y.~Li, J.~Wu, and L.~Fei-Fei, ``Voxposer:
  Composable 3d value maps for robotic manipulation with language models,''
  \emph{arXiv preprint arXiv:2307.05973}, 2023.

\bibitem{kerr2023lerf}
J.~Kerr, C.~M. Kim, K.~Goldberg, A.~Kanazawa, and M.~Tancik, ``Lerf: Language
  embedded radiance fields,'' in \emph{Proceedings of the IEEE/CVF
  International Conference on Computer Vision}, 2023, pp. 19\,729--19\,739.

\bibitem{OpenScene}
S.~Peng, K.~Genova, C.~M. Jiang, A.~Tagliasacchi, M.~Pollefeys, and
  T.~Funkhouser, ``Openscene: 3d scene understanding with open vocabularies,''
  2023.

\bibitem{conceptfusion}
K.~M. Jatavallabhula, A.~Kuwajerwala, Q.~Gu, M.~Omama, T.~Chen, S.~Li, G.~Iyer,
  S.~Saryazdi, N.~Keetha, A.~Tewari \emph{et~al.}, ``Conceptfusion: Open-set
  multimodal 3d mapping,'' 2023.

\bibitem{engelmann2024opennerf}
F.~Engelmann, F.~Manhardt, M.~Niemeyer, K.~Tateno, M.~Pollefeys, and
  F.~Tombari, ``{OpenNeRF: Open Set 3D Neural Scene Segmentation with
  Pixel-Wise Features and Rendered Novel Views},'' in \emph{International
  Conference on Learning Representations}, 2024.

\bibitem{kim2024garfield}
C.~M. Kim, M.~Wu, J.~Kerr, K.~Goldberg, M.~Tancik, and A.~Kanazawa, ``Garfield:
  Group anything with radiance fields,'' in \emph{Proceedings of the IEEE/CVF
  Conference on Computer Vision and Pattern Recognition}, 2024, pp.
  21\,530--21\,539.

\bibitem{mccormac2018fusion++}
J.~McCormac, R.~Clark, M.~Bloesch, A.~Davison, and S.~Leutenegger, ``Fusion++:
  Volumetric object-level slam,'' in \emph{2018 international conference on 3D
  vision (3DV)}.\hskip 1em plus 0.5em minus 0.4em\relax IEEE, 2018, pp. 32--41.

\bibitem{zhang2020flowfusion}
T.~Zhang, H.~Zhang, Y.~Li, Y.~Nakamura, and L.~Zhang, ``Flowfusion: Dynamic
  dense rgb-d slam based on optical flow,'' in \emph{2020 IEEE international
  conference on robotics and automation (ICRA)}.\hskip 1em plus 0.5em minus
  0.4em\relax IEEE, 2020, pp. 7322--7328.

\bibitem{newcombe2015dynamicfusion}
R.~A. Newcombe, D.~Fox, and S.~M. Seitz, ``Dynamicfusion: Reconstruction and
  tracking of non-rigid scenes in real-time,'' in \emph{Proceedings of the IEEE
  conference on computer vision and pattern recognition}, 2015, pp. 343--352.

\bibitem{luiten2023dynamic}
J.~Luiten, G.~Kopanas, B.~Leibe, and D.~Ramanan, ``Dynamic 3d gaussians:
  Tracking by persistent dynamic view synthesis,'' \emph{arXiv preprint
  arXiv:2308.09713}, 2023.

\bibitem{xie2024physgaussian}
T.~Xie, Z.~Zong, Y.~Qiu, X.~Li, Y.~Feng, Y.~Yang, and C.~Jiang, ``Physgaussian:
  Physics-integrated 3d gaussians for generative dynamics,'' in
  \emph{Proceedings of the IEEE/CVF Conference on Computer Vision and Pattern
  Recognition}, 2024, pp. 4389--4398.

\bibitem{fu2022robust}
J.~Fu, Y.~Du, K.~Singh, J.~B. Tenenbaum, and J.~J. Leonard, ``Robust change
  detection based on neural descriptor fields,'' in \emph{2022 IEEE/RSJ
  International Conference on Intelligent Robots and Systems (IROS)}.\hskip 1em
  plus 0.5em minus 0.4em\relax IEEE, 2022, pp. 2817--2824.

\bibitem{qiu2020indoor}
Y.~Qiu, Y.~Satoh, R.~Suzuki, K.~Iwata, and H.~Kataoka, ``Indoor scene change
  captioning based on multimodality data,'' \emph{Sensors}, vol.~20, no.~17, p.
  4761, 2020.

\bibitem{looper20233d}
S.~Looper, J.~Rodriguez-Puigvert, R.~Siegwart, C.~Cadena, and L.~Schmid, ``3d
  vsg: Long-term semantic scene change prediction through 3d variable scene
  graphs,'' in \emph{2023 IEEE International Conference on Robotics and
  Automation (ICRA)}.\hskip 1em plus 0.5em minus 0.4em\relax IEEE, 2023, pp.
  8179--8186.

\bibitem{Schmid-RSS24-Khronos}
L.~Schmid, M.~Abate, Y.~Chang, and L.~Carlone, ``Khronos: A unified approach
  for spatio-temporal metric-semantic slam in dynamic environments,'' in
  \emph{Proc. of Robotics: Science and Systems (RSS)}, Delft, Netherlands, July
  2024.

\bibitem{qi2017pointnet}
C.~R. Qi, H.~Su, K.~Mo, and L.~J. Guibas, ``Pointnet: Deep learning on point
  sets for 3d classification and segmentation,'' in \emph{Proceedings of the
  IEEE conference on computer vision and pattern recognition}, 2017, pp.
  652--660.

\bibitem{wang2019dgcnn}
Y.~Wang, Y.~Sun, Z.~Liu, S.~E. Sarma, M.~M. Bronstein, and J.~M. Solomon,
  ``Dynamic graph cnn for learning on point clouds,'' \emph{ACM Transactions on
  Graphics (tog)}, vol.~38, no.~5, pp. 1--12, 2019.

\bibitem{park2019deepsdf}
J.~J. Park, P.~Florence, J.~Straub, R.~Newcombe, and S.~Lovegrove, ``Deepsdf:
  Learning continuous signed distance functions for shape representation,'' in
  \emph{Proceedings of the IEEE/CVF conference on computer vision and pattern
  recognition}, 2019, pp. 165--174.

\bibitem{chen2019learning}
Z.~Chen and H.~Zhang, ``Learning implicit fields for generative shape
  modeling,'' in \emph{Proceedings of the IEEE/CVF conference on computer
  vision and pattern recognition}, 2019, pp. 5939--5948.

\bibitem{lin2023coarse}
C.-W. Lin, T.-I. Chen, H.-Y. Lee, W.-C. Chen, and W.~H. Hsu, ``Coarse-to-fine
  point cloud registration with se (3)-equivariant representations,'' in
  \emph{2023 IEEE international conference on robotics and automation
  (ICRA)}.\hskip 1em plus 0.5em minus 0.4em\relax IEEE, 2023, pp. 2833--2840.

\bibitem{simeonov2022neural}
A.~Simeonov, Y.~Du, A.~Tagliasacchi, J.~B. Tenenbaum, A.~Rodriguez, P.~Agrawal,
  and V.~Sitzmann, ``Neural descriptor fields: Se (3)-equivariant object
  representations for manipulation,'' in \emph{2022 International Conference on
  Robotics and Automation (ICRA)}.\hskip 1em plus 0.5em minus 0.4em\relax IEEE,
  2022, pp. 6394--6400.

\bibitem{fu2023neuse}
J.~Fu, Y.~Du, K.~Singh, J.~B. Tenenbaum, and J.~J. Leonard, ``Neuse: Neural se
  (3)-equivariant embedding for consistent spatial understanding with
  objects,'' in \emph{Proceedings of Robotics: Science and Systems (RSS)},
  2023.

\bibitem{chen2020simple}
T.~Chen, S.~Kornblith, M.~Norouzi, and G.~Hinton, ``A simple framework for
  contrastive learning of visual representations,'' in \emph{International
  conference on machine learning}.\hskip 1em plus 0.5em minus 0.4em\relax PMLR,
  2020, pp. 1597--1607.

\bibitem{qi2017pointnet++}
C.~R. Qi, L.~Yi, H.~Su, and L.~J. Guibas, ``Pointnet++: Deep hierarchical
  feature learning on point sets in a metric space,'' \emph{Advances in neural
  information processing systems}, vol.~30, 2017.

\bibitem{sohn2016improved}
K.~Sohn, ``Improved deep metric learning with multi-class n-pair loss
  objective,'' \emph{Advances in neural information processing systems},
  vol.~29, 2016.

\bibitem{oquab2023dinov2}
M.~Oquab, T.~Darcet, T.~Moutakanni, H.~Vo, M.~Szafraniec, V.~Khalidov,
  P.~Fernandez, D.~Haziza, F.~Massa, A.~El-Nouby \emph{et~al.}, ``Dinov2:
  Learning robust visual features without supervision,'' \emph{arXiv preprint
  arXiv:2304.07193}, 2023.

\bibitem{li2022lseg}
\BIBentryALTinterwordspacing
B.~Li, K.~Q. Weinberger, S.~Belongie, V.~Koltun, and R.~Ranftl,
  ``Language-driven semantic segmentation,'' in \emph{International Conference
  on Learning Representations}, 2022. [Online]. Available:
  \url{https://openreview.net/forum?id=RriDjddCLN}
\BIBentrySTDinterwordspacing

\bibitem{ester1996density}
M.~Ester, H.-P. Kriegel, J.~Sander, and X.~Xu, ``Density-based spatial
  clustering of applications with noise,'' in \emph{Int. Conf. knowledge
  discovery and data mining}, vol. 240, no.~6, 1996.

\bibitem{ravi2024sam2}
N.~Ravi, V.~Gabeur, Y.-T. Hu, R.~Hu, C.~Ryali, T.~Ma, H.~Khedr, R.~R{\"a}dle,
  C.~Rolland, L.~Gustafson \emph{et~al.}, ``Sam 2: Segment anything in images
  and videos,'' \emph{arXiv preprint arXiv:2408.00714}, 2024.

\bibitem{teed2021droid}
Z.~Teed and J.~Deng, ``Droid-slam: Deep visual slam for monocular, stereo, and
  rgb-d cameras,'' \emph{Advances in neural information processing systems},
  vol.~34, pp. 16\,558--16\,569, 2021.

\end{thebibliography}

\end{document}